# ECOC-Based Training of Neural Networks for Face Recognition

Nima Hatami
Department of Electrical Engineering
Shahed University
Tehran, Iran
hatami@shahed.ac.ir

Reza Ebrahimpour
School of Cognitive Sciences, Institute for Studies on Theoretical Physics and Mathematics (IPM)
Tehran, Iran
ebrahimpour@ipm.ir

Reza Ghaderi
Department of Electrical Engineering
Mazandaran University
Babol, Iran
r_ghaderi@nit.ac.ir

*Abstract*— Error Correcting Output Codes, ECOC, is an output representation method capable of discovering some of the errors produced in classification tasks. This paper describes the application of ECOC to the training of feed forward neural networks, FFNN, for improving the overall accuracy of classification systems. Indeed, to improve the generalization of FFNN classifiers, this paper proposes an ECOC-Based training method for Neural Networks that use ECOC as the output representation, and adopts the traditional Back-Propagation algorithm, BP, to adjust weights of the network. Experimental results for face recognition problem on Yale database demonstrate the effectiveness of our method. With a rejection scheme defined by a simple robustness rate, high reliability is achieved in this application.

## I. Introduction

Multi-class classifiers have wide and practical usages in pattern recognition for problems that involve several possible categories. Given a training sample vector $x=\{x_1, \ldots, x_l\}$, where $l$ is the dimension of the data sample, the task of a multiclass classifier is to assign it to one of the $C$ categories with $C \geq 3$. Examples of such applications include in optical digit recognition ($C = 10$); diagnosis of different diseases based on medical signals and face recognition problem.

The standard neural network approach to multiclass problems is to construct a 3-layer feed forward network with $C$ output units, where each output unit designates one of the $C$ classes. During training, the output units are clamped to 0, except for the unit corresponding to the desired class, which is clamped at 1. During classification, a new $x$ value is assigned to the class whose output unit has the highest activation. Let us call this the one-per-class approach.

Ref [1, 2] showed that an alternative method, called error-correcting output coding (ECOC) gives superior performance. In this approach, each class $i$ is assigned a $b$-bit binary string, $c_i$, called a codeword. The strings are chosen so that the Hamming distance between each pair of strings is guaranteed to be at least d. During training on example x, the b output units of a 3-layer network are clamped to the appropriate binary string $c_i$. During classification, the new example x is assigned to the class $i$ whose codeword $c_i$ is closest, in Hamming distance, to the $b$-element vector of output activations. The advantage of this approach is that it can recover from any $[d-1/2]$ errors in learning the individual output units.

There are two main approaches to the design of a classifier using OC methods, depending on the characteristics of the decomposition unit:

• Monolithic classifier unit which is composed of a monolithic classifier with multiple outputs, exploiting the decomposition in an implicit way. Examples are multiple-input multiple-output (MIMO) learning machines, such as MIMO MLP (Multi-layer Perceptron) or MIMO decision trees [1, 2].

• Parallel classifiers unit is implemented by an ensemble of dichotomizers, assigning each dichotomy to a different dichotomizer. Consequently, the learning task is distributed among separated and independent dichotomizers, each learning a different bit of the codeword coding a class [3, 4].

Consequently, there are three problems, when using ECOC in neural networks (like the monolithic MLP with BP learning rule):

1- As in Ref [5] concluded, the BP algorithm is not able to recover error-correcting output codes by itself. This gives additional evidence that ECOC provides an additional source of power for improving neural network generalization performance. Along with their results, this suggests that error-correcting output codes should be adopted (instead of the one-per-class approach) as the standard method for applying BP to multiclass problems.

2- In many case, for achieving satisfactory results, we tend to increase length of codeword. This leads to a network with a large number of output nodes, called output complexity. Large networks tend to introduce high internal interference because of the strong coupling among their hidden-layer weights [6]. Internal interference exists during the training process; when updating the weights of hidden units the influence (desired outputs) from two or more output units cause the weights to compromise to nonoptimal values due to the clash in their weight update directions. Therefore, we should modify BP algorithm to overcome this shortcoming.

3- As mentioned in ref [2], the individual bits of error-correcting codes are much more difficult to learn than the bits in the one-per-class approach or the Sejnowski-Rosenberg distributed code. This leads to the networks with higher complexities and more hidden neurons needed to handle the task, which make the learning process (finding optimal weights and parameters) more difficult. Therefore, we must try to modify neural network learning rules to be adapted to the ECOC representations.

To address these issues, in this paper, we introduce a modified version of the BP algorithm to improve the performance of the monolithic-ECOC MLP network. In our method, we use the function of error, produced on over all codeword, as the weights in the cost function term for a more efficient updating of the network weights.

The paper is divided into the following sections: Section 2 provides a brief introduction to error correcting output codes, Section 3 describes the proposed method to make the BP algorithm compatible with the ECOC technique and computational modeling of face recognition. Section 4 shows experimental results of the proposed method and Section 5 concludes the paper.

## II. ERROR-CORRECTING OUTPUT CODES IN MULTI-CLASS CLASSIFICATION

Original idea of ECOC is motivated from signal transmission in communication which class information is transmitted over a noisy channel. In this method using codes with error correcting properties, we propose strategy to suppress existing noise effects. In classification problems, this noise is dichotomies' error caused by limited training samples, complexity of class boundary, over fitting of base classifiers or any misclassification factors and using ECOC in classification leads to overcome this shortcoming and increase generalization.

The ECOC algorithm for the monolithic classifier can be reviewed as follows:

### ECOC Algorithm

*Training phase:*

For each $C \times b$ code matrix, ($C$ is the number of classes)

- Codify label of each class with rows of the code matrix.

- Train monolithic classifier with the patterns based on new defined labels.

Therefore, we have a classifier with $b$ output nodes

*Test phase:*

- Apply an incoming test pattern $x$ to the trained classifier and create an output vector:

$$\overline{y} = [y_1, y_2, ..., y_b]^T \qquad (1)$$

where $y_j$ is the output of $j$th output node.

For decision making (reconstruction)

- For each class, measure distance between the output vector and label of each class (matrix row):

$$L_i = \sum_{j=1}^{b} |Z_{ij} - y_j| \qquad (2)$$

Where $Z_{ij}$ is a member of $i$th row and $j$th column in code matrix.

- Assign $x$ to the class $c_j$ corresponding to the closest code word:

$$i = ArgMin(L_i) \qquad (3)$$

We face three main problems in designing monolithic-ECOC classifiers:

• Code generation methods for effective decomposition: various methods have been proposed in the literature to code matrix generation [1, 7]. The BCH code generation, exhaustive codes, randomized hill climbing are some famous code generation methods with good results in the literature [1, 7, 8]. In almost all of these methods, the final goal is to have greatest possible distance between any pair of code words for more error-correcting capability of the network. Recently, some code generation methods have been proposed which consider problem structure [14, 15]. This leads to more efficient decomposition of the problem.

• Preparing suitable network architecture and learning rule: since the overall classification accuracy highly depends on network architecture and its adjusted weights; we should try to choose the best architecture and learning rules compatible with ECOC algorithm.

• Appropriate reconstruction strategy design: Many different strategies proposed in reconstruction stage such as minimum distance, dempster-shafer based combining, the least squares method and the centroid algorithm [8, 1, 7]. In all of these methods, an incoming pattern is assigned to a class according to closest distance to a binary code word (row of matrix).

## III. COMPUTATIONAL MODELLING OF FACE RECOGNITION

The model consists of two processing stages: representation and recognition (Fig. 1). In the representation stage, any input retinal image is transformed to a low dimension vector, appropriate representation in the MLP input. The recognition stage, which is of vital importance, is an ECOC monolithic MLP with the proposed improved BP learning algorithm. The next subsections describe the two processing stages of the model in more detail.

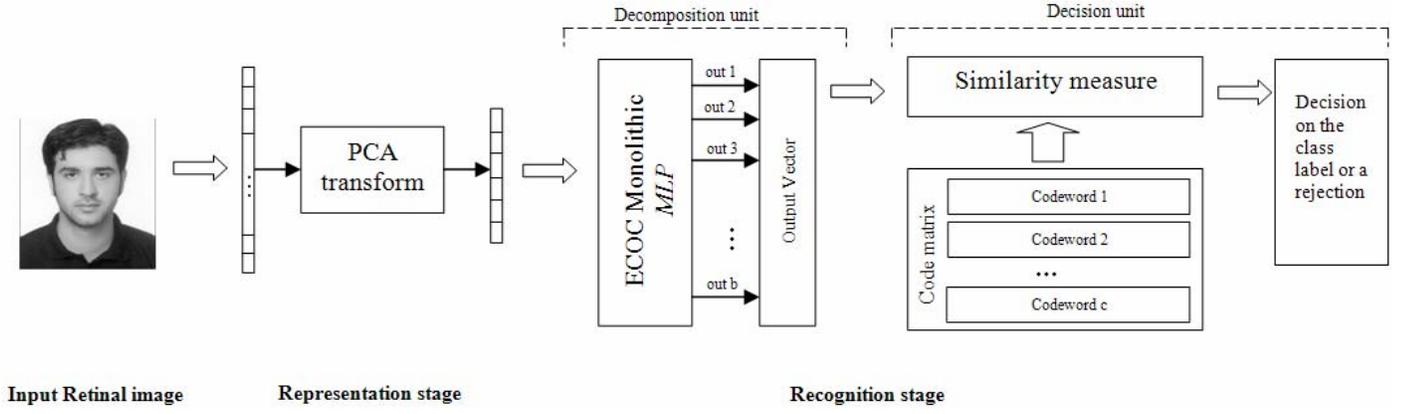

Fig. 1 The proposed model consists of two main stages: face representation and recognition.

*Representation Stage:*

In the first stage of our face recognition model, we use PCA, principal component analysis, to avoid a high dimensional and redundant input space, and optimally design and train the binary classifiers. The resulting low-dimensional representation is used for face processing. PCA is the simplest and most efficient method for coding faces [9], however other methods such as linear discriminant analysis, LDA, [10, 11] and independent-component analysis, ICA, [12] have revealed good results. For the current model, it is only important to have a low dimensional code, to ease generalization to new faces.

PCA method is implemented in the following steps: after normalizing the data, calculate its covariance matrix; third calculating the eigenvectors and eigenvalues of the covariance matrix, and then choosing appropriate components and forming a feature vector.

*Recognition Stage:*

We used the ECOC monolithic MLP in the recognition stage of our model. Here, we introduce a modified BP algorithm, which is more adapted to ECOC method. The standard feed forward neural networks learning rules use one-per-class codes for output representation. Their training problems are generally posed in terms of unconstrained optimization where the objective is to minimize a squared-error cost function of the form

$$E(w) = \frac{1}{N}\sum_{i=1}^{N}(y(w,u_i)-d_i)^2 \quad (4)$$

Defined on the training set

$$\tau = \{u_i, d_i\}, \quad i \in 1,...,N \quad (5)$$

with respect to the network parameters $w$. Here $N$ is the size of the training set and $u_i, d_i$ and $y(w,u_i)$ are the input vector, desired network output and actual network output for the $i$th training vector, respectively.

$$w_{k+1} = w_k + \eta_k f(g_k) \quad (6)$$

where

$$g = \left[\frac{\partial E(w_1)}{\partial w_1} \frac{\partial E(w_2)}{\partial w_2} \cdots \frac{\partial E(w_b)}{\partial w_b}\right]^T \quad (7)$$

Here $g_k$ is the gradient of the cost function and $\eta_k$ is the step size at the $k$th iteration. The descent direction computation, $f(.)$, dictates the rate of convergence achievable with each method and is typically a trade-off between performance and computational/memory requirements.

As mentioned before, any incoming sample to ECOC is misclassified, if the total error introduced by its codeword is longer than $\left[d-\frac{1}{2}\right]$. Our approach, introduces a weight to adjust the importance of error in the output node produced by each sample. This way, when introduced error on target codeword is high (resulting in misclassification); produced error in each output nodes has more effect in the training cost function. This leads to efficient updating the weight parameters of network and results in error reduction on total codeword by trained network. The modified cost function ($\bar{E}$) is then in the form of Eq. (8)

$$\bar{E}(w) = \frac{1}{N}\sum_{i=1}^{N}\omega_i(y(w,u_i)-d_i)^2 \quad (8)$$

where $\omega_i$ is the weight for the error produced by $i$th sample, total summation of errors produced in target codeword by $i$th sample, and is defined by Eq. (9).

$$\omega_i = \sum_{j=1}^{b}(y_{ij}(w,u_i)-d_{ij})^2 \quad (9)$$

where $j$ and $b$ are the number of output nodes and length of codeword respectively.

Here we extend the idea and define the robustness rate of a decision of the proposed face recognition model as follows:

$$RR = \frac{Hd(cw_2, \bar{y}) - Hd(cw_1, \bar{y})}{Hd(cw_2, cw_1)} \times 100 \quad (10)$$

Where $cw_1$ and $cw_2$ are the closest and second closest rows of the code matrix to the output vector, $\bar{y}$, given by ECOC classifier for each test sample and $Hd(.,.)$ is the Hamming distance between two codewords. A robustness threshold can be set on RR so that testing samples with RR smaller than the threshold can be rejected. The threshold can be adjusted based on tradeoff between recognition rate and error rate. Finally, the reliability of the face recognition model defined as follows:

$$Reliability = \frac{Recognition\ Rate}{Recognition\ Rate + Error\ Rate} \quad (11)$$

## IV. EXPERIMENTAL RESULTS

The Yale face database is used in our experiments. This database contains 165 gray scale images of 15 individuals, 11 images for each individual. The images demonstrate variations in lighting condition, facial expression (normal, happy, sad, sleepy, surprised, and wink) and accessories. Samples of the Yale face database are shown in Fig. 2.

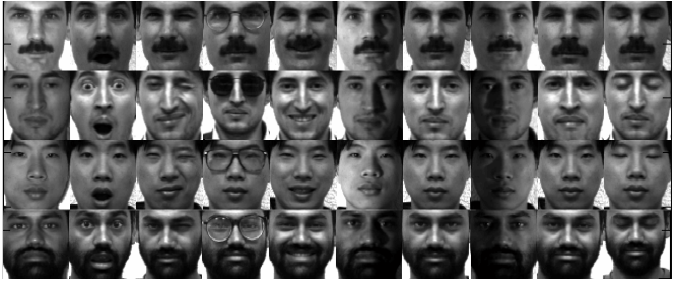

Fig. 2 Example of face images, with variation in pose, facial expression, and illumination.

All the face images are manually aligned and cropped. The size of each cropped image is 32 × 32 pixels, with 256 gray levels per pixel. The features, pixel values, are then scaled to [0,1] (divided by 256). For the vector-based approaches, the image is represented as a 1024-dimensional vector, which by using the PCA transform, a feature vector was created by the 30 largest PCA values [13].

The image set is then partitioned into the training and test set with different numbers. For easier representation, *Gm/Pn* means *m* images per person are randomly selected for training and the remaining *n* images are for testing. The recognition accuracy of standard BP and proposed algorithms on Yale is reported on the Table (1). For each *Gm/Pn*, we average the results over 10 random splits. In this experiment, we used the BCH code with the size of 15×31, as a code matrix. We compare the performance of the proposed method with the traditional BP algorithm with the same structure and parameters. As shown in table 1, our proposed method demonstrated better performance in terms of higher recognition rate in comparison with the BP algorithm. However, recognition rate in the both networks increases, as the number of training samples increases.

TABLE I. RECOGNITION ACCURACY ON YALE FOR DIFFERENT NUMBER OF TRAINING AND TESTING SAMPLES (%)

| Method | G2/P9 | G4/P7 | G6/P5 | G8/P3 |
|---|---|---|---|---|
| Standard BP | 52.29 | 59.8 | 70.21 | 77.77 |
| Proposed method | 53.18 | 61.7 | 71.1 | 78.51 |

In another experiment, we investigate the effect of different code matrixes on the recognition rates of the two compared algorithms. We used 15×63 BCH code, 15×105 one vs. one, 15×15 one vs. all, 15×59 Sparse random code and 15×39 Dense random code matrix; and *G6/P5* for training/testing of the networks. As shown in table 2, our method outperforms the BP algorithm. However, one vs. one code has the longer codewords, which results in higher number of output nodes and more difficult weights adjustment, but it helps the networks achieve higher accuracy.

TABLE II. RECOGNITION ACCURACY ON YALE FOR DIFFERENT CODE MATRIX (%)

| Method | BCH-15 | BCH-31 | BCH-63 | 1vs.1 | 1vs. All | Sparse random | Dense random |
|---|---|---|---|---|---|---|---|
| Standard BP | 70.44 | 70.21 | 68.66 | 71 | 69.59 | 70 | 65.33 |
| Proposed method | 71.1 | 71.1 | 68.98 | 71.9 | 69.86 | 70.5 | 66.13 |

We then make use of the robustness rate defined in Eq. (10) to generate rejections and minimize the error rate of proposed face recognition model. The robustness threshold is set equal to 25%. We used *G8/P3* for training/testing of the networks. Table 3 gives the Reliability of the proposed model on Yale database for different code matrixes. From the results, we observe that our approach is capable of achieving promising results on the face recognition task.

TABLE III. RELIABILITY OF THE PROPOSED FACE RECOGNITION MODEL ON YALE DATABASE FOR DIFFERENT CODE MATRIXES (%).

| Method | BCH-15 | BCH-31 | BCH-63 | 1vs.1 | 1vs. A | Sparse random | Dense random |
|---|---|---|---|---|---|---|---|
| Reliability | 90.6 | 89.5 | 89.8 | 91.3 | 88 | 89.8 | 85.5 |

## V. CONCLUSION

In this paper, we introduced an improved BP algorithm for the training neural network with ECOC output representation. For this purpose, we introduced a weight factor in the cost function to adjust the importance of introduced error by each training sample. This weight is adjusted in a way that the total error obtained by target codeword, which cause misclassification in ECOC classifiers, is reduced. We validated our proposed method with the face recognition problem on the Yale database. Experimental results for different size of training/testing sets and code matrixes show the robustness of our proposed method but further analyses are needed to investigate the origin of its efficiency.